# Liberating language research from dogmas of the 20th century

*Ramon Ferrer-i-Cancho[1] & Carlos Gómez-Rodríguez[2]*

**Abstract.** A commentary on the article "Large-scale evidence of dependency length minimization in 37 languages" by Futrell, Mahowald & Gibson (PNAS 2015 112 (33) 10336-10341).

Central to the inspiring contributions of E. Gibson and collaborators to language research is the idea that a wide range of phenomena, e.g., ambiguity resolution, parsing difficulties or even our notion of sentence "grammaticality", could be manifestations of a principle of dependency length minimization (e.g., references to Gibson's work of Futrell, Mahowald, & Gibson, 2015), in stark contrast to the view of generative linguistics at least.

In a recent study of impressive breadth, Futrell, Mahowald and Gibson (2015) have provided evidence of dependency length minimization across languages by means of various baselines. Paradoxically, the random baselines incorporate constraints on word order that are likely to be consequences of the very principle of dependency length minimization. Futrell et al. argue that their "Free Word Order Baseline" does not obey any particular word order rule, however, it is not actually free because crossing dependencies are not allowed. A truly free word order baseline, and indeed a fully null hypothesis, is one where the $n!$ possible linearizations of the $n$ units (words) of a sentence are allowed *a priori*, as in the pioneering research on dependency length minimization by Ferrer-i-Cancho that Futrell et al. (2015) cite. Furthermore, a large body of theoretical and empirical research strongly suggests that non-crossing dependencies arise as a side effect of pressure to reduce dependency lengths (see Gómez-Rodríguez & Ferrer-i-Cancho 2016, Ferrer-i-Cancho & Gómez-Rodríguez, 2015 and references therein).

Therefore, investigating dependency length minimization with random baselines or an "optimal baseline" where crossings are not allowed is not only theoretically superficial, but also unnecessarily complicated and most worryingly, indicates subordination to the division between competence and performance, a dogma of generative linguistics that Gibson and collaborators have challenged in the past. Futrell et al.'s "Consistent Head Direction Baseline" is another example of baseline that is likely to incorporate dependency length minimization in its very definition: consistent head direction might be a consequence of dependency length minimization (Ferrer-i-Cancho, 2015a, 2015b). For instance, once the verb is placed last (as in SOV order), dependency length minimization predicts that, consistently, the dependents of the nominal heads of S and O should precede their heads. Similar arguments can be made for the "Fixed word order baseline": dependency length minimization predicts the relative placements for certain dependencies, e.g. adjectives with respect to their nominal heads, verbal auxiliaries with respect to their verbal heads, and so on (Ferrer-i-Cancho, 2015a).

[1] Complexity & Qualitative Linguistics Lab, LARCA Research Group, Departament de Ciències de la Computació, Universitat Politècnica de Catalunya, Campus Nord, Edifici Omega. Jordi Girona Salgado 1-3. 08034 Barcelona, Catalonia, Spain.
Address correspondence to: rferrericancho@cs.upc.edu.
[2] LyS Research Group, Departamento de Computación, Facultade de Informática, Universidade da Coruña, Campus de A Coruña, 15071 A Coruña, Spain.

Surprisingly, Futrell et al. take for granted dogmas behind principles and parameters theory, where the consistent branching is assumed (not explained) and its direction is determined by a parameter. In contrast, tendencies for consistent branching and its direction are less parameter-consuming predictions of a mathematical theory of dependency length minimization (Ferrer-i-Cancho, 2015a, 2015b).

In sum, Futrell et al.'s research on dependency length minimization is an example of radical empirical research that attempts to remain theoretically agnostic but, paradoxically, turns out to gullibly accept tenets of theoretical linguistics of the past century. Those tenets can be summarized as a belief in the existence of word order constraints that cannot be explained by evolutionary processes or requirements of performance or learning, and instead require either (a) heavy assumptions that compromise the parsimony of linguistic theory as a whole or (b) explanations based on internal constraints of obscure nature.

Our commentary has focused on the problems of Futrell et al.'s analysis for the construction of a general theory of language that is both highly predictive and parsimonius. Other issues have been reviewed by Liu, Xu, and Liang (2016).

**Acknowledgments**

This commentary is a slightly extended version of the letter that we submitted to PNAS and was rejected. R.F.C is funded by the grants 2014SGR 890 (MACDA) from AGAUR (Generalitat de Catalunya) and the grant TIN201457226-P from MINECO (Ministerio de Economia y Competitividad). C.G.R is partially funded by the MINECO grant FFI2014-51978-C2-2-R and Xunta de Galicia (grant R2014/034 and an Oportunius program grant).